\documentclass{article}
\usepackage{spconf,amsmath,graphicx,hyperref}
\usepackage{booktabs}
\usepackage{multirow}
\usepackage{makecell}
\usepackage{enumitem}
\usepackage{wasysym}   %\Letter

\title{Following the TRACE: A Structured Path to Empathetic Response Generation with Multi-Agent Models}
\name{Ziqi Liu, Ziyang Zhou, Yilin Li, Haiyang Zhang, Yangbin Chen$^{\dagger}$\thanks{$^{\dagger}$ Corresponding author. Email: Yangbin.Chen@xjtlu.edu.cn}}
\address{School of Advanced Technology (SAT)\\
Xi'an Jiaotong-Liverpool University, Suzhou, China
}

\begin{document}
%\ninept
%
\maketitle
\begin{abstract}
Empathetic response generation is a crucial task for creating more human-like and supportive conversational agents. However, existing methods face a core trade-off between the analytical depth of specialized models and the generative fluency of Large Language Models (LLMs). To address this, we propose \textbf{TRACE}, \textbf{T}ask-decomposed \textbf{R}easoning for \textbf{A}ffective \textbf{C}ommunication and \textbf{E}mpathy, a novel framework that models empathy as a structured cognitive process by decomposing the task into a pipeline for analysis and synthesis. By building a comprehensive understanding before generation, TRACE unites deep analysis with expressive generation. Experimental results show that our framework significantly outperforms strong baselines in both automatic and LLM-based evaluations, confirming that our structured decomposition is a promising paradigm for creating more capable and interpretable empathetic agents. Our code is available at \url{https://github.com/sunbus100/TRACE}.
\end{abstract}
\section{INTRODUCTION}

Enhancing human-computer interaction with the ability to understand and respond to human emotions is a crucial frontier in artificial intelligence, with significant applications in areas such as mental health support and automated customer service. However, generating truly empathetic responses is more than just fluent text generation. It requires modeling a complex, multi-level cognitive process that encompasses accurate affective perception, deep causal reasoning, and appropriate strategic planning \cite{wang2022care,fu2023reasoning}.

Research in empathetic response generation has historically followed two main paradigms. Traditional specialized models excelled at the analytical aspects, leveraging knowledge graphs or commonsense reasoning to achieve a deep understanding of specific emotional contexts \cite{li2022knowledge,sabour2022}. However, they were often limited by their model scale, struggling to produce responses with high linguistic fluency and diversity. The advent of LLMs has revolutionized generative capabilities, enabling the creation of highly natural and coherent responses. This has led to a core trade-off in the current landscape between analytical depth and generative power. While expressive, LLMs often lack a deep, structured understanding of the user's state, which can lead to responses that are generic, superficial, and fail to provide targeted empathy \cite{chen2024dialoguellm}.

To address this trade-off, we argue for a paradigm that explicitly unites structured analysis with powerful generation. In this paper, we propose TRACE, a novel multi-agent framework that models empathy as a structured, multi-stage cognitive process. Our framework decomposes the task into a four-stage pipeline handled by specialized agents: 1. emotion recognition, 2. causal analysis, 3.strategic planning, and 4. response synthesis. By building a comprehensive, layer-by-layer understanding of the user's state before generation, TRACE aims to produce responses that are both deeply understanding and highly expressive.

The main contributions of this work are as follows:
\begin{itemize}[noitemsep] 
    \item We propose a novel multi-agent framework that combines the strengths of deep analysis and fluent generation by decomposing the empathetic process into a structured pipeline. 
    \item We demonstrate the effectiveness of our framework through extensive experiments, showing that it significantly outperforms strong baselines \cite{fu2023reasoning,chen2024dialoguellm}. 
    \item Our framework provides inherent interpretability by design, as its pipeline architecture explicitly models the reasoning path from emotion and cause analysis to the final communication strategy.
\end{itemize}

\section{RELATED WORK}
\subsection{Modeling Empathetic Understanding}
Early research in Empathetic Response Generation primarily focused on modeling the user's inner state to build a foundation for empathetic interaction \cite{rashkin2019, chen2022wish}. This pursuit branched into two main research streams, affective perception and cognitive reasoning. Efforts in affective perception concentrated on emotion recognition, with models developed for both coarse-grained, utterance-level categories \cite{lin2019, majumder2020} and more nuanced, word-level emotional cues \cite{gao2021improving, li2022knowledge}. In parallel, the cognitive reasoning stream aimed to deepen a model's comprehension of the user's situation, often by integrating external knowledge sources like commonsense graphs \cite{sabour2022, cai-etal-2024-empcrl}. While these specialized approaches advanced the analytical aspects of empathy, a persistent limitation was the difficulty in generating responses that matched the sophistication of their analytical insights, often due to constrained model parameters \cite{bi2023diffusemp}.

\subsection{Generative Fluency and The Synthesis Challenge}
The advent of LLMs such as ChatGPT \cite{achiam2023} marked a significant leap in generative fluency for ERG. By leveraging vast pre-training, LLMs excel at producing coherent and contextually appropriate language, overcoming the expressive limitations of earlier models \cite{zhou2022large, zhao2023}. However, this fluency has not always been accompanied by a corresponding depth of empathetic understanding. Multiple studies indicate that LLMs can falter in fine-grained emotional perception and may not perform the necessary reasoning to uncover the root causes of a user's feelings \cite{liu2025caf}. This highlights a trade-off between analytical depth and generative power. We propose TRACE to address this, a multi-agent framework that decomposes the empathetic process into a structured pipeline. By dedicating distinct agents to emotion perception, causal analysis, and strategic planning, our approach is designed to unite the deep, structured understanding of specialized models with the powerful, expressive generation of LLMs.

\section{METHODOLOGY}
\subsection{Overall Pipeline}
\begin{figure}[!htp]
    \centering
    \includegraphics[width=0.49\textwidth]{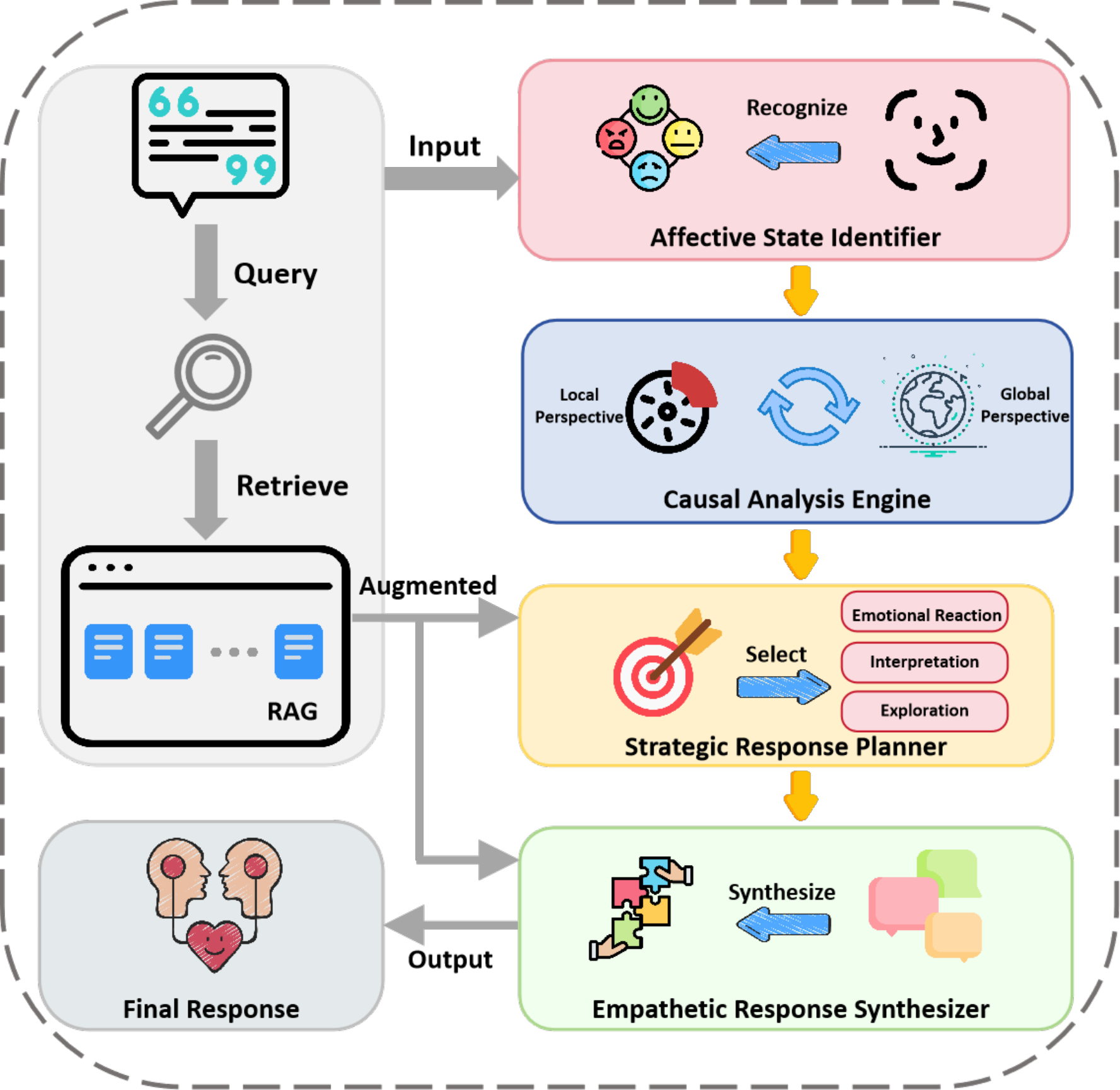}
    
    \caption{The overall architecture of our TRACE framework. The system processes a dialogue history through a four-agent pipeline, where each agent enriches the context with a specific layer of analysis before the final response is synthesized. Agents 3 and 4 leverage a RAG system to consult a knowledge base of examples.}
    \label{fig:framework}
\end{figure}

To generate nuanced empathetic responses, we propose a multi-agent framework that decomposes this complex task into a structured, multi-stage pipeline. The framework sequentially processes a user's dialogue, allowing for the progressive enrichment of context to ensure a deeply grounded response. The pipeline begins with the \textbf{Affective State Identifier (ASI)}, which first perceives the user's core emotion. Next, the \textbf{Causal Analysis Engine (CAE)} analyzes the specific reasons for this emotional state. This output then informs the \textbf{Strategic Response Planner (SRP)}, which selects an optimal communicative strategy. Finally, the \textbf{Empathetic Response Synthesizer (ERS)} integrates this entire chain of analysis to generate a final, high-quality, and context-aware reply. The overall architecture of this pipeline is depicted in Figure \ref{fig:framework}.

\subsection{Affective State Identifier}
The ASI agent is responsible for the foundational task of accurately perceiving the seeker's emotional state from the dialogue history $D$. Human emotion is inherently complex; however, to create a computationally tractable and psychologically grounded framework, we adopt Paul Ekman's influential theory of basic emotions \cite{ekman1992argument}. Specifically, we map the 32 fine-grained emotion categories from the original dataset into the six coarse-grained, universal emotions. This simplification yields a more robust classification task and ensures the resulting emotional context is both meaningful and interpretable for subsequent agents. Let this set of emotions be $\mathcal{E}$. The classification process is formalized as:
\begin{equation} \label{eq:emotion_classification}
e^* = \underset{e \in \mathcal{E}}{\operatorname{argmax}} P(e | D)
\end{equation}
The agent's output is a single emotion label $e^*$, which provides the core emotional context for all subsequent agents.

\subsection{Causal Analysis Engine}
Following emotion identification, the CAE agent is invoked to diagnose the underlying reasons for the identified emotion $e^*$. This agent performs a dual-granularity analysis to build a comprehensive understanding. The entire process can be viewed as a function $f_{CAE}$ that maps the dialogue and emotion to a structured analysis object $O_{CAE}$:
\begin{equation} \label{eq:cae_func}
O_{CAE} = f_{CAE}(D, e^*, prompt)
\end{equation}
Specifically, this function comprises two sub-tasks. From a local perspective, a function $f_{trigger}$ identifies the key conversational turns that serve as trigger spans $S_{trigger}$. From a global perspective, a function $f_{global}$ formulates a global cause summary $c_{sum}$ and, drawing from emotional evaluation theories, assigns a psychological cause category $c_{cat}$ to the situation \cite{ortony1990cognitive}. This categorization is based on a taxonomy of common emotion-eliciting situations. This can be represented as:
\begin{equation} \label{eq:trigger_func}
S_{trigger} = f_ {trigger}(D, e^*, prompt)
\end{equation}
\begin{equation} \label{eq:global_func}
(c_{sum}, c_{cat}) = f_{global}(D, e^*, prompt)
\end{equation}
The final output is $O_{CAE} = (S_{trigger}, c_{sum}, c_{cat})$, which provides a rich causal analysis containing these distinct analytical components.

\subsection{Strategic Response Planner}
As the core decision-making component, the SRP agent selects a communicative strategy from a predefined set $\mathcal{S}$, which includes \textbf{Emotional Reaction (ER)}, \textbf{Interpretation (IP)}, and \textbf{Exploration (EX)}. To ground its decision in successful precedents, the agent leverages a RAG system. This system retrieves relevant dialogue sessions from our training set $\mathcal{C}$, by comparing their initial scenario prompts. These scenario prompts are short situations provided by the researcher that establish the premise for each conversation. Relevance is quantified by a semantic similarity score $\sigma$, calculated between the embedding vector of the current dialogue scenario prompt $p$ and that of each candidate session $p_i$:
\begin{equation} \label{eq:similarity}
\sigma(p, p_i) = \frac{\mathbf{v}_p \cdot \mathbf{v}_{p_i}}{\|\mathbf{v}_p\| \|\mathbf{v}_{p_i}\|}
\end{equation}
The RAG system first conducts a Precise Search for examples where $e_i = e^*$ and $\sigma(p, p_i) > \tau$. If none are found, a Fuzzy Search is performed, relaxing the condition to $\sigma(p, p_i) > \tau$ alone. The optimal strategy $s^*$ is then selected based on all prior analysis and the retrieved examples $\mathcal{K}$:
\begin{equation} \label{eq:strategy_selection}
s^* = f_{strat}(D, e^*, O_{CAE}, \mathcal{K})
\end{equation}

\subsection{Empathetic Response Synthesizer}
The final agent, ERS, generates the empathetic reply $R$. It synthesizes the comprehensive analytical context ($e^*$, $O_{CAE}$, $s^*$) from the preceding agents to ensure the response is coherent and strategically targeted. Furthermore, to align the output with effective conversational patterns, the ERS also employs a RAG system analogous to the one in the SRP. This system retrieves stylistic exemplars, allowing the ERS to generate a final reply that is not only analytically grounded but also stylistically mirrors proven empathetic responses by mimicking their tone and phrasing. This generation process is represented by:
\begin{equation} 
R = G_{resp}(D, e^*, O_{CAE}, s^*, \mathcal{K}')
\end{equation}

\section{EXPERIMENTS}
\subsection{Experimental Setup}

\textbf{Datasets:} We conduct experiments on the widely-used ED dataset \cite{rashkin2019}, following the standard data splits for our evaluation.

\textbf{Baselines:} We compare our framework against two categories of baselines. 
\textbf{1. Specialized Models:} We select a comprehensive set of SOTA models including Multi-TRS \cite{rashkin2019}, EmpDG \cite{li2020}, KEMP \cite{li2022knowledge}, CEM \cite{sabour2022}, CASE \cite{zhou2022case}, and EmpSOA \cite{zhao2023}.
\textbf{2. PLM-based Models:} We also compare against pre-trained dialogue models, including BlenderBot \cite{roller2021}, DialoGPT \cite{zhang2020}, and LEMPEx \cite{majumder2022exemplars}. \textbf{3. LLM-based Methods:} We also include methods based on LLMs, such as EmpGPT-3 \cite{lee2022does} and EmpCRL \cite{cai-etal-2024-empcrl}. The performance results for all baseline models are reported directly from the EmpCRL \cite{cai-etal-2024-empcrl} to ensure a fair comparison.

\textbf{Implementation Details:} Our framework is implemented via the GPT-4o API. To ensure the reproducibility of the analysis stages, the first three agents operate with a deterministic temperature setting of 0. The final ERS agent uses a temperature of 0.5 to encourage response diversity.

\textbf{Evaluation Metrics:} \textbf{1. Automatic Evaluation:} Given the poor correlation of reference-based metrics with human judgment in dialogue \cite{liu2016}, our evaluation instead focuses on intrinsic qualities. We measure fluency via Perplexity \cite{serban2015}; diversity using Distinct-n \cite{li2016} and EAD-n \cite{liu2022rethinking}; and emotional understanding via Emotion Accuracy (I-ACC) \cite{singh2021fine}. \textbf{2. Human-like Evaluation:} We use GPT-4o as an automated assessor, a method shown to highly correlate with human judgment. On 100 random samples, GPT-4o conducts a pairwise A/B test, comparing TRACE against our baseline on Empathy, Relevance, and Fluency to produce Win/Lose/Tie statistics.

\begin{table*}[!htp]
\centering
\caption{Results of automatic evaluation. The best results among all models are highlighted in \textbf{bold}, second-best are \underline{underlined}.}
\label{tab:auto_eval_results_final}
{ % 开始一个分组，让设置仅在此表格内生效
\renewcommand{\arraystretch}{0.8} % 调整行高 (小于1为缩小)
\begin{tabular*}{\textwidth}{@{\extracolsep{\fill}} llcccccc}
\toprule
\textbf{Type} & \textbf{Models} & \textbf{PPL}& \textbf{Dist-1} & \textbf{Dist-2} & \textbf{EAD-1} & \textbf{EAD-2} & \textbf{I-ACC} \\
\midrule
\multirow{6}{*}{\textit{Transformer-based}} 
& Multi-TRS & 39.15 & 0.32 & 1.24 & 0.96 & 2.87 & 20.06 \\
& EmpDG & 36.45 & 0.47 & 1.89 & 1.41 & 3.97 & 24.51 \\
& KEMP & 37.96 & 0.51 & 2.12 & 1.09 & 3.48 & 29.15 \\
& CEM & 37.47 & 0.65 & 2.76 & 1.13 & 3.69 & 26.81 \\
& CASE & 35.79 & 0.71 & 3.85 & 1.47 & 4.96 & 32.41 \\
& EmpSOA & 35.98 & 0.65 & 3.51 & 1.44 & 4.21 & 30.99 \\
\midrule
\multirow{3}{*}{\textit{PLM-based}} 
& LEMPEx & 26.37 & 1.41 & 14.66 & 3.51 & 13.85 & - \\
& BlenderBot & \underline{16.71} & 2.58 & 11.55 & 2.24 & 16.80 & - \\
& DialoGPT & 18.74 & 2.71 & 12.01 & 2.87 & 16.51 & - \\
\midrule
\multirow{2}{*}{\textit{LLM-based}} 
& EmpGPT-3 & - & 3.15 & \underline{18.63} & 4.25 & 17.50 & - \\
& EmpCRL & \textbf{15.70} & \underline{4.27} & 16.11 & \underline{5.39} & \underline{22.63} & \underline{41.57} \\
& \textbf{Ours} & 18.35 & \textbf{13.62} &\textbf{48.12} &\textbf{10.26} &\textbf{50.20} & \textbf{44.28} \\
\bottomrule
\end{tabular*}
} 
\end{table*}

\begin{table}[!htp]
\centering
\caption{LLM-based evaluation results. Each comparison was conducted three times, and all reported improvements of our model are statistically significant ($p < 0.05$).}
\label{tab:llm_eval_results}
{ % 开始一个分组，让设置仅在此表格内生效
\renewcommand{\arraystretch}{0.9} % 您可以微调这个值来控制行高
% 使用 tabular* 和 \columnwidth 来自动填充单栏宽度
\begin{tabular*}{\columnwidth}{@{\extracolsep{\fill}} llcc}
\toprule
\textbf{Comparisons} & \textbf{Aspects} & \textbf{Win} & \textbf{Lose} \\
\midrule
% --- Comparison against the backbone model ---
\multirow{4}{*}{\textbf{vs. GPT-4o}}
& Empathy (Emp.)       & 80\% & 20\% \\
& Informativity (Inf.) & 74\% & 26\% \\
& Fluency (Flu.)       & 79\% & 21\% \\
& Consistency (Con.)   & 85\% & 15\% \\
\midrule
% --- Comparison against the SOTA baseline ---
\multirow{4}{*}{\textbf{vs. EmpGPT-3}}
& Empathy (Emp.)       & 58\% & 42\% \\
& Informativity (Inf.) & 63\% & 37\% \\
& Fluency (Flu.)       & 61\% & 39\% \\
& Consistency (Con.)   & 65\% & 35\% \\
\bottomrule
\end{tabular*}
} % 结束分组
\end{table}

\subsection{Main Result}
\subsubsection{Automatic Evaluation Results}
The automatic evaluation results in Table \ref{tab:auto_eval_results_final} demonstrate the effectiveness of our framework, establishes new state-of-the-art results on key metrics while remaining highly competitive on others, showcasing a strong overall capability.

TRACE exhibits superior performance in generation diversity, dramatically surpassing all baselines on both Distinct-n and EAD-n. This success is attributable to our multi-agent pipeline, which enables the final agent to focus on generating creative responses. Furthermore, our framework achieves the highest I-ACC of 44.28, validating the efficacy of our analytical agents in accurately perceiving the user's emotion.

Regarding fluency, TRACE achieves a competitive PPL score, indicating that its superior diversity and accuracy do not compromise response coherence. Overall, these results confirm that our structured, multi-agent approach successfully unites deep empathetic understanding with expressive and diverse generation.

\subsubsection{Human-like Evaluation Results}

The results of our LLM-based evaluation are presented in Table \ref{tab:llm_eval_results}. In the critical comparison against its own backbone model, our framework, TRACE, significantly outperforms a directly prompted GPT-4o across all four evaluated aspects. Notably, it achieves its largest win margins in Empathy with a 80\% win rate and in Consistency with a 85\% win rate. This finding is crucial, as it provides strong evidence that our structured multi-agent pipeline adds significant value beyond the raw generative capability of the underlying LLM, validating our decompositional approach.

Furthermore, when benchmarked against EmpGPT-3, a strong state-of-the-art baseline, TRACE continues to demonstrate superior performance across all criteria, confirming its effectiveness within the current research landscape. Collectively, these evaluations support our core hypothesis: by modeling empathy as a structured pipeline of analysis encompassing emotion, cause, and strategy, our framework produces responses that are consistently judged to be more empathetic, informative, fluent, and consistent.

\subsection{Ablation Experiment}
\begin{table}[!htp]
\centering
\caption{Ablation study results on diversity metrics. The best results are highlighted in \textbf{bold}.}
\label{tab:ablation_study}
\renewcommand{\arraystretch}{0.9}
\begin{tabular}{lcccc}
\toprule
\textbf{Model Variant} & \textbf{Dist-1} & \textbf{Dist-2} & \textbf{EAD-1}& \textbf{EAD-2}\\
\midrule
\textbf{Full Model} & \textbf{13.62} &\textbf{48.12} &\textbf{10.26} &\textbf{50.20} \\
\midrule
w/o RAG & 9.76 & 39.01 & 8.70 & 48.57 \\
w/o ASI & 12.98 & 46.05 & 9.96 & 49.10 \\
w/o CAE & 13.28 & 46.43 & 9.91 & 48.89 \\
w/o SRP & 10.20 & 40.08 & 8.12 & 42.18 \\
\bottomrule
\end{tabular}
\end{table}

We conducted an ablation study to verify the contribution of each component in the TRACE framework, with results on diversity metrics shown in Table \ref{tab:ablation_study}. The results clearly indicate that each component positively contributes to the final performance. The w/o Analysis Pipeline and w/o RAG variants show the most substantial degradation in all diversity scores, which validates that our structured analysis and the retrieval of exemplars are crucial for generating diverse responses. Furthermore, the w/o ASI, w/o CAE, and w/o SRP variants also lead to a noticeable, albeit smaller, performance drop, confirming that each layer of analysis incrementally contributes to the richness of the final output.

\section{CONCLUSION}

In this paper, we proposed TRACE, a novel multi-agent framework that addresses the trade-off between analytical depth and generative fluency by decomposing the empathetic process into a structured pipeline of specialized agents. Experiments show that TRACE significantly outperforms strong baselines in diversity and emotional accuracy, while LLM-based evaluations confirm its superior empathetic quality. We conclude that this structured decomposition is a promising paradigm for creating empathetic agents that are both deeply understanding and highly expressive.

\newpage
\footnotesize
\section{Acknowledgment}
This research was supported by the National Natural Science Foundation of China (Grant No.~62502396) and the XJTLU Research Development Fund (RDF-24-02-008).

\footnotesize
\bibliographystyle{IEEEbib}
\bibliography{strings,refs}

\end{document}